%% file: main.tex
\documentclass{article}

\PassOptionsToPackage{numbers, compress}{natbib}

     \usepackage[final]{neurips_2019}

\usepackage[utf8]{inputenc} 
\usepackage[T1]{fontenc}    
\usepackage{hyperref}       
\usepackage{url}            
\usepackage{booktabs}       
\usepackage{amsfonts}       
\usepackage{nicefrac}       
\usepackage{microtype}      
\usepackage{color}
\usepackage{natbib}
\usepackage{graphicx}
\usepackage{bbm}
\usepackage{listings}

\definecolor{dkgreen}{rgb}{0,0.6,0}
\definecolor{gray}{rgb}{0.5,0.5,0.5}
\definecolor{mauve}{rgb}{0.58,0,0.82}
\begin{document}

\title{VALAN: Vision and Language Agent Navigation\thanks{Code: \url{https://github.com/google-research/valan}}}

\author{
Larry Lansing \\
Google Research \\
{\tt \small{ llansing@google.com}} \\
\And
Vihan Jain \\
Google Research \\
{\tt \small{ vihanjain@google.com}} \\
\And
Harsh Mehta \\
Google Research \\
{\tt  \small{harshm@google.com}} \\
\And
Haoshuo Huang\thanks{Work done as an intern at Google Research.}\\
Carnegie Mellon University\\
{\tt  \small{haoshuoh@andrew.cmu.edu}} \\
\And
Eugene Ie\\
Google Research \\
{\tt  \small{eugeneie@google.com}} \\
}


\maketitle

\begin{abstract}
VALAN is a lightweight and scalable software framework for deep reinforcement learning based on the SEED RL architecture. The framework facilitates the development and evaluation of embodied agents for solving grounded language understanding tasks, such as Vision-and-Language Navigation and Vision-and-Dialog Navigation, in photo-realistic environments, such as Matterport3D~\cite{Chang2017Matterport3DLF} and Google StreetView~\cite{Chen19:touchdown}. We have added a minimal set of abstractions on top of SEED RL allowing us to generalize the architecture to solve a variety of other RL problems. In this article, we will describe VALAN's software abstraction and architecture, and also present an example of using VALAN to design agents for instruction-conditioned indoor navigation.
\end{abstract}

\input{introduction}

\input{design}

\input{case_study_r2r}
%
%

\section{Acknowledgements}
We thank Alexander Ku, Austin Waters, Gabriel Ilharco, Jason Baldridge and Ming Zhao for their valuable comments and inputs that helped shaped VALAN.

\bibliographystyle{plain}

\input{main.bbl}
\end{document}

%% file: introduction.tex
\section{Introduction}

The ability to train on massive amounts of data is crucial to achieving state-of-the-art results in machine learning. This dependence on data is even more pronounced in deep reinforcement learning (RL) given sample complexity concerns. Vision-and-Language Navigation (VLN)~\cite{Anderson:2018:VLN} and Vision-and-Dialog (VDN) \cite{thomason:corl19} are complex tasks for studying embodied agent designs, where agents must interpret natural language to achieve prescribed navigation goals in photo-realistic environments. These agents learn navigation policies using deep RL in cross-modal contexts further pushing the data scaling requirements necessary for competitive end-to-end solutions.

Solving VLN/VDN problems effectively requires the use of modeling architectures that scale horizontally. One such scalable learning architecture is SEED RL~\cite{espeholt2019seed}.  It addresses scalability challenges by distributing expensive simulations across large numbers of actor machines and centralizes the model training in a single learner that combines the distributed actor roll-outs using an off-policy correction algorithm called V-trace~\cite{espeholt:impala}. This setup is particularly attractive for VLN/VDN as the expensive simulation of navigation in photo-realistic environments can be fanned out across machines, opening up opportunities to perform more expensive end-to-end deep training/inference of vision and language modules for better domain adaptation. 

VALAN implements a variant of SEED RL/IMPALA~\cite{espeholt:impala} to address the scalability challenges inherent in VLN/VDN datasets. As of writing, the approach that provides the best performance involves a mix of supervised learning and policy gradients, an alternating teacher/student-forcing training regime. VALAN supports both training methodologies making it complete for further experimentation. One other key addition is the ability to do per-episode computation once and re-use that for every state in the episode. This is critical for VLN/VDN use cases where the instruction typically needs to be encoded once per episode to avoid duplicating that computation for every state. This gets prohibitively expensive as the episode length increases.

In this short article, we will outline the overall design of VALAN and use a case study to demonstrate how to build agents for the popular Room-to-Room instruction-conditioned indoor navigation task. Prototypes for other embodied agent design problems are available on our Github repository.

\paragraph{Related Packages} There are many packages available specifically for RL research and development. For example, Dopamine~\cite{DBLP:journals/corr/abs-1812-06110} and TF-agents~\cite{TFAgents} both contain a comprehensive suite of RL algorithms with interfaces supplied for evaluating them on standard RL problems in popular environments such as OpenAI Gym~\cite{OpenAI_Gym:arxiv16} and ALE~\cite{bellemare:jair2013}. Our package focuses on a specific distributed RL architecture, namely SEED RL  because of its ease to scale processing across many actor machines, to study computationally intensive multi-modal fusion and control problems for embodied agent design.


%% file: design.tex
\section{Design}

\begin{figure*}
  \centering
  \includegraphics[width=\textwidth]{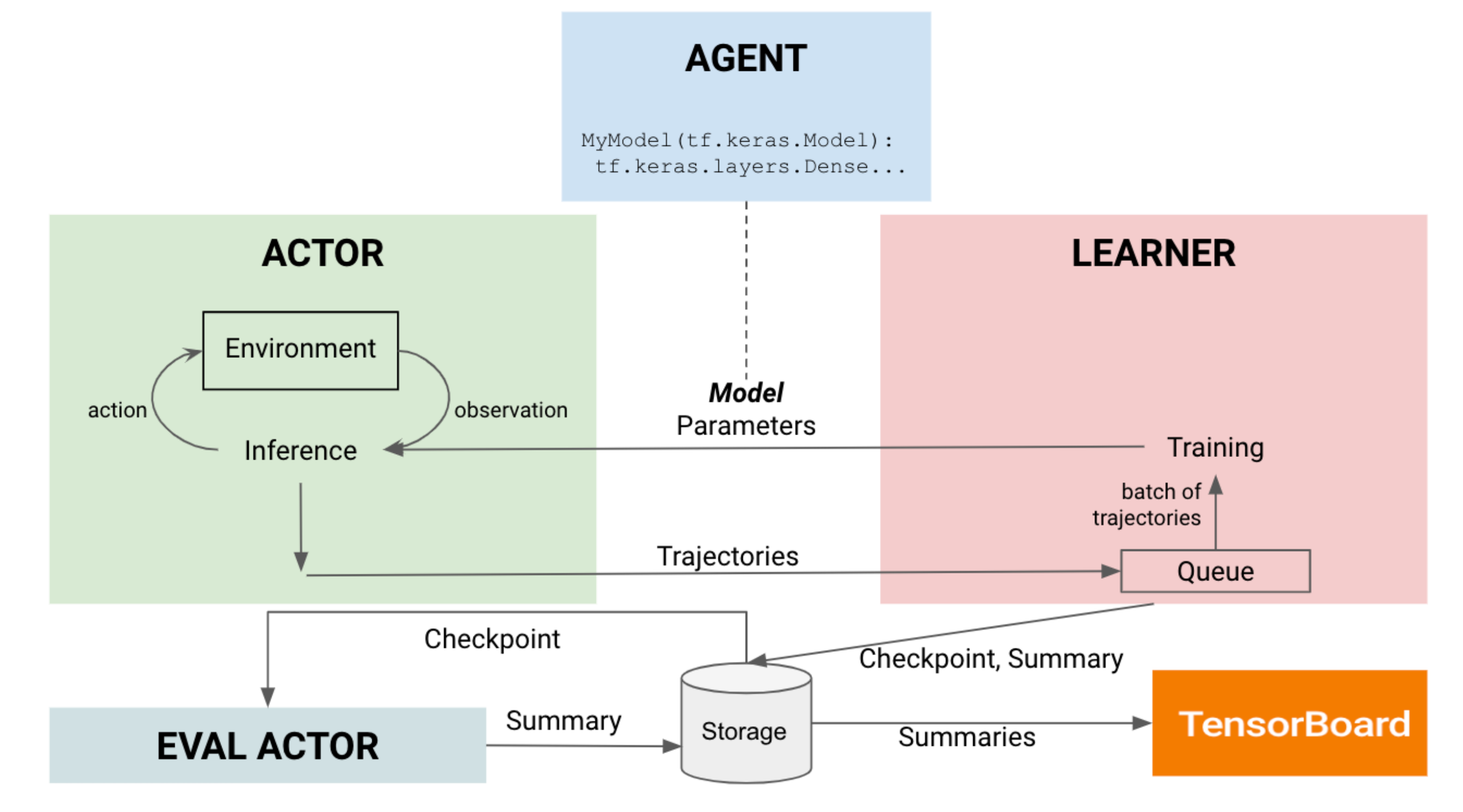}
  \caption{Main design elements in VALAN\label{fig:design_elements}}
\end{figure*}

VALAN is implemented using TensorFlow 2.0~\cite{abadi2016tensorflow} and is comprised of five main design elements: Agent, Environment, Actor, Learner and Problem. Figure~\ref{fig:design_elements} depicts the interaction between them.

\subsection{Agent}

The Agent is a TensorFlow \lstinline{tf.keras.Model} implementation of the machine learning model being trained to solve the current Problem. The Agent model is executed by both the Actor and the Learner in inference and learning modes respectively. Researchers can instantiate multiple Agent implementations in the same Environment to evaluate various modeling strategies.

\begin{figure*}
  \centering
  \includegraphics[width=\textwidth]{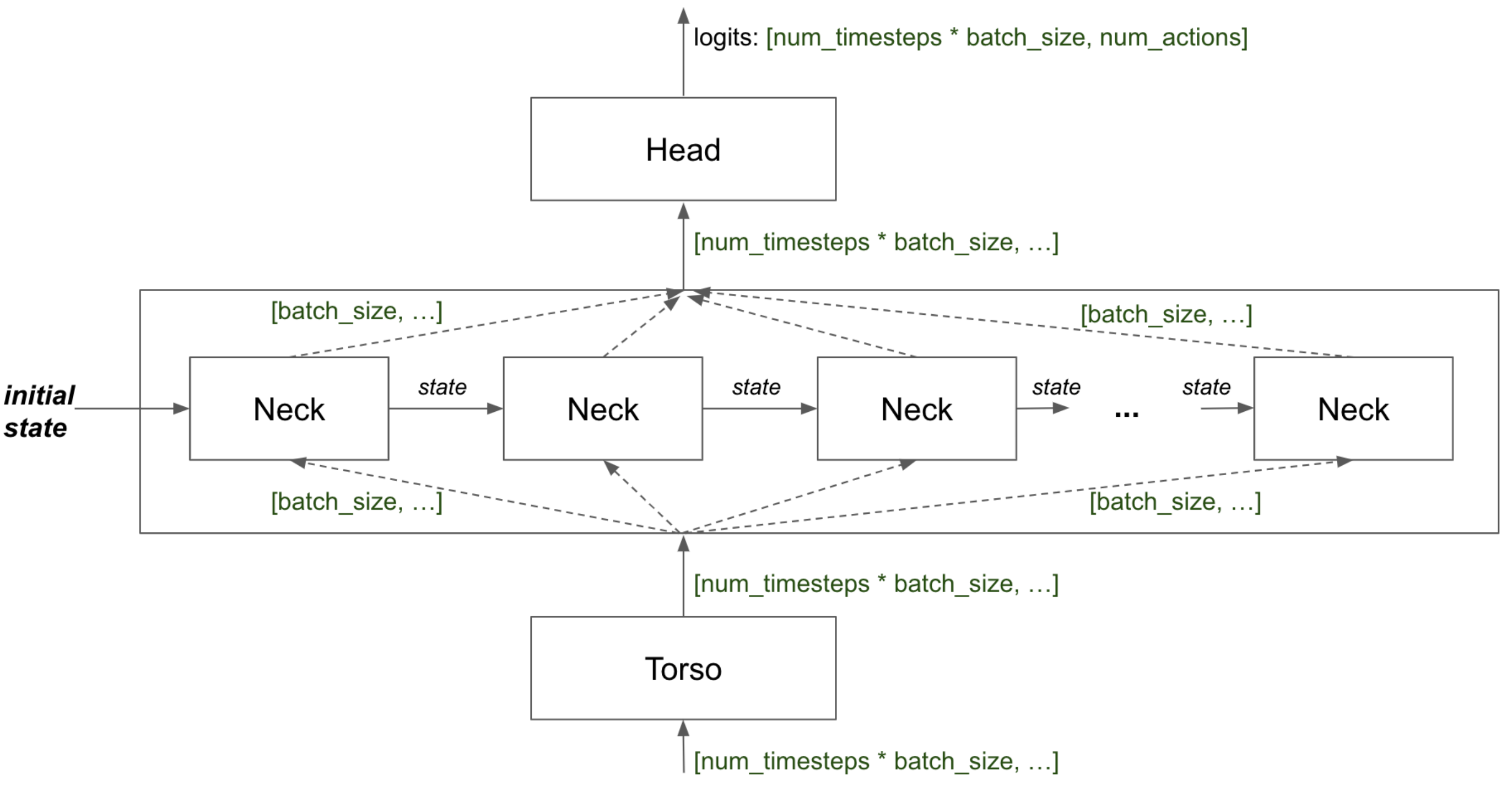}
  \caption{Data flow in core components of Agent\label{fig:agent_data_flow}}
\end{figure*}

Recurrent Neural Nets \cite{Rumelhart:1988:LRB:65669.104451} are frequently used in VLN agents to retain historical contexts and actions for next action prediction. This architectural choice requires Agents to generate actions one step at a time. However, there are often computationally intensive initializations required at each step of the trajectory (e.g. encoding the full instruction text). There are also additional post-processing computations that are performed once per trajectory after the last step executes. VALAN's Agent API supports this standard operating procedure by splitting the Agent into three components as shown in Figure~\ref{fig:agent_data_flow}.

\begin{itemize}
    \item \textbf{Torso}: The Torso contains pre-processing logic that is executed once per trajectory before the first recurrent step is taken, with output applied to all time steps in that trajectory. For example, if the input at every time step is a per step image (which is passed through several convolutional layers), the images for all time steps can be processed together as a batch. The first dimension of all inputs to Torso is \lstinline{[num_timesteps * batch_size]}.
    \item \textbf{Neck}: The Neck contains per step logic that depends on the recurrent state from the previous time step. This is called once per time step in a trajectory, and each call also receives the output from the Torso. The first dimension of inputs to Neck is \lstinline{[batch_size]}.
    \item \textbf{Head}: The Head contains post-processing logic that is executed once per trajectory after the last step is taken. It is given the outputs of each call to Neck across all the time steps. The first dimension of inputs to Head is \lstinline{[num_timesteps * batch_size]}.
\end{itemize}



\subsection{Environment}

The Environment defines the valid state transitions and reward function for a given problem. It tracks changes to the agent’s state as it steps through an episode. The interface is similar to the OpenAI Gym environment interface\footnote[1]{http://gym.openai.com/docs/\#environments}. All agents navigate the same Environment for a given problem. It is important to note that the Environment module is encapsulated by the Actor allowing the practitioner to sample observations based on different policy roll-outs in different environments. By distributing the execution of different policies across machines, the learner can sample the observations for a given policy across different partitioning schemes (e.g. partition machines by different indoor houses, cities; or partition machines by training curricula). 

At every step, the Environment outputs a \lstinline{EnvOutput} tuple which has the following values:

\begin{itemize}
    \item \textbf{Observation}: An object or dictionary of objects representing the current observation of the Environment, e.g. current visual scene, current Agent’s heading.
    \item \textbf{Reward}: A float value indicating the reward achieved by the previous action.
    \item \textbf{Done}: A boolean value indicating whether the Environment state should be resetted. If true, it signifies the end of the current episode (e.g., the Agent predicted STOP action or the Agent exhausted its budget of steps).
    \item \textbf{Info}: A dictionary containing any diagnostic information useful for debugging.
\end{itemize}

\subsection{Actor}

As highlighted above, the Actor is a module that interacts with the Environment to obtain the current time step's observation (as well as the reward obtained from executing the Agent's action to previous time step's input) and infers the action using agent's current policy. The actor generates a trajectory of such steps until either the agent predicts a special action denoting the end of the episode (STOP action) or the agent's budget is exhausted. The generated trajectories are sent to the Learner which can employ various learning algorithms to update agent's policy. The Actor periodically copies the latest Agent model weights from the Learner for use in inference. Thousands of Actors may run in parallel, each sending trajectories to the same Learner.

The actors can run either in training mode where they send the generated trajectories to learner or in evaluation mode (referred to as \textit{eval actor}) where the generated trajectories along with the computed validation metrics are sent to an aggregator which aggregates metrics across multiple trajectories and displays them on user interfaces such as command line or dashboards like Tensorboard.

\subsection{Learner}

The Learner is a program that trains on trajectories from Actors to learn agent policies that satisfies the given objective (e.g., maximizing the rewards or maximizing log-likelihood of expert demonstration actions). The learner receives trajectories from multiple Actors in parallel, processing them as a batch in order to make optimal use of hardware accelerators such as GPUs and TPUs~\cite{google-tpu}. The learner is a typical ML model orchestration code where the batches are dequeued from a queue, tunable hyperparameters are supported (e.g., learning schedule, gradient clipping) and optional training summaries are printed on user dashboards.

\subsection{Problem}
The Problem is a Python class that contains problem-specific code such as validation metrics, training optimizer and loss functions.

%% file: case_study_r2r.tex
\section{Case Study}
Since the primary motivation for building VALAN is for solving problems involving multiple input modalities such as VLN, we will now describe the framework's key components in the context of experimenting with the Room-to-Room navigation problem. The readers may find it useful to peruse the actual code available on Github\footnote[2]{https://github.com/google-research/valan/tree/master/r2r} while reading the following subsections to understand how a problem can be implemented on VALAN from scratch.

\subsection{The Room-to-Room Dataset}
The Room-to-Room (R2R) dataset~\cite{DBLP:journals/corr/abs-1807-06757} is based on 90 houses from the Matterport3D environments~\cite{Chang2017Matterport3DLF} each defined by an undirected graph. The nodes are locations where egocentric photo-realistic panoramic images are captured and the edges define the connections between locations. The dataset consists of language instructions paired with reference paths, where each path is a sequence of graph nodes. Each path is associated with 3 natural language instructions collected using Amazon Mechanical Turk with an average token length of 29 from a dictionary of 3.1k unique words. Paths collected are longer than 5m and contain 4 to 6 edges. The dataset is split into a training set, two validation sets and a test set. One validation set includes new instructions on environments overlapping with the training set (Validation Seen), and the other is entirely disjoint from the training set (Validation Unseen). 

Evaluation on the validation unseen set and the test set assess the agent's full generalization ability. Metrics for assessing agents performance include: \textit{Navigation Error (NE)} which is a measure of the distance between agent's stopping location and the goal; \textit{Success Rate (SR)} measures how often the agent stops within some acceptable distance of the goal; \textit{Success weighted by Path Length (SPL)}~\cite{DBLP:journals/corr/abs-1807-06757} measures whether the SR success criteria was met, weighted by the normalized path length; \textit{Coverage weighted by Length Score (CLS)}~\cite{jain2019stay} measures the agent path’s conformity to the ground truth path weighted by length score; and \textit{Success weighted by normalized Dynamic Time Warping (SDTW)}~\cite{magalhaes2019effective} is \textit{SR} weighted by DTW score of agent's path.

\subsection{VALAN Environment for R2R}
Writing environment implementation for any problem involves inheriting from the VALAN \lstinline{BaseEnv} class and defining the two abstract methods: \lstinline{reset} and \lstinline{step}. Here we assume that a collection of paths along with their associated natural language instructions are cached in the class object.

The \lstinline{reset} method is called at the beginning of a new agent trajectory. The output of this method is \lstinline{EnvOutput} tuple that contains: reward at the beginning of the agent's trajectory (which is usually zero), the boolean `done' set to false that indicates this is not the end of agent's trajectory, the `observation' dictionary contains tensors that will be used in the ML model in the agent (see Sec.~\ref{subsec:agent-r2r}) and `info' contains debugging information used for providing statistics about learning progress. In the R2R environment, the main tensors used by the agent are the encoding of the current visual scene (e.g., last layer's output of a pre-trained CNN model), the instructions for the current path (which can be tokenized) and the encoding of the neighboring states that the agent may choose to visit next.

The \lstinline{step} method is called at every step of the agent's trajectory. The output of this method is the same \lstinline{EnvOutput} tuple. The main difference from \lstinline{reset} method is that in this method, the reward needs to be computed for the current step and it also needs to be evaluated if the agent is at the end of its current trajectory (e.g., it ran out of fixed budget of steps or it predicted STOP action).

\subsection{VALAN Agent for R2R}
\label{subsec:agent-r2r}
The agent is the TensorFlow model used for learning the navigation policy in learner and for inference to create new trajectories in actor. In the open-sourced VALAN repository, we provided the implementation of RCM agent~\cite{wang2019reinforced} which has the following salient characteristics:

\begin{itemize}
    \item \textit{{get\_initial\_state}}: This is called once at the beginning of a trajectory. Since the instruction for a trajectory is fixed, we execute the instruction encoder (which is a bidirectional LSTM~\cite{Schuster1997BidirectionalRN}) in this method.
    \item \textit{torso}: This method runs no TensorFlow ops in the implementation.
    \item \textit{neck}: In this method, which is called once every step in the trajectory, we run the visual encoder (which is an LSTM) that encodes the current step's visual scene. The output of LSTM is then used as query to attend over the instruction encoder output computed in \textit{{get\_initial\_state}}.
    \item \textit{head}: In this method, we compute attention over all the visual scenes of the current trajectory using the output of \textit{neck} as the query. Finally, we compute the probability distribution over agent's neighbor nodes at each of the steps in the trajectory.
\end{itemize}

The R2R implementation uses two types of losses --- policy gradient loss computed using rewards obtained by executing agent's own policy as well as supervised learning loss for trajectories where the agent is forced to mimic actions from the logs (expert demonstrations) and the objective is to increase the maximum likelihood of choosing the correct neighbor node at every step.

\begin{figure*}
  \centering
  \includegraphics[width=\textwidth]{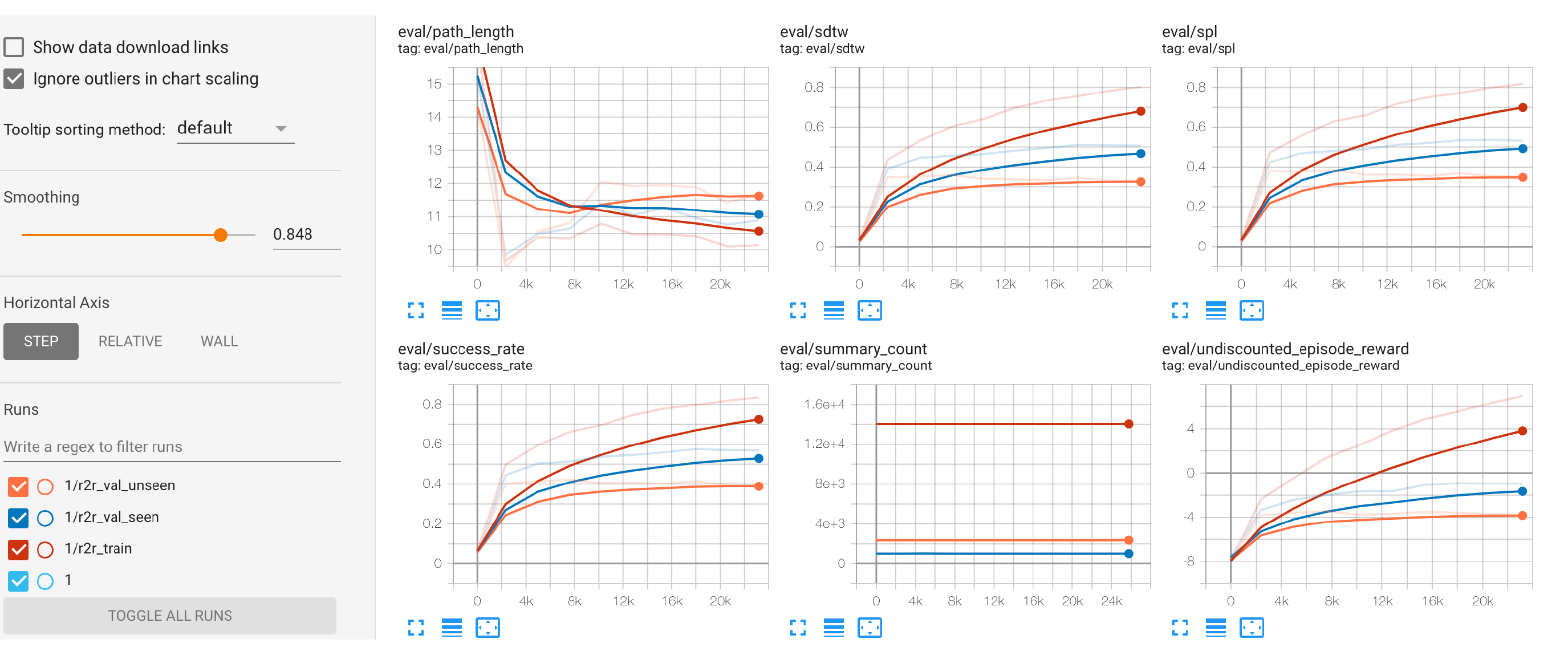}
  \caption{Tensorboard displaying validation metrics computed for an RCM agent trained on R2R dataset. \label{fig:r2r_case_study}}
\end{figure*}

\subsection{VALAN Problem for R2R}
The problem definition contains methods that define the current problem's characteristics like the environment, agent and optimizer to be used, the training summaries (for debugging) and evaluation summaries (SR, NE, SPL, SDTW etc.) to present at the dashboards. The following methods are noteworthy for R2R implementation:

\begin{itemize}
    \item \textit{{get\_episode\_loss\_type}}: As mentioned above, we use two different types of losses for different trajectories while learning. This method chooses which loss type to use for the given trajectory.
    \item \textit{{select\_actor\_action}}: This method chooses if the actor should sample action from agent's policy (suitable for trajectories with policy gradient loss) or from available demonstration trajectories (suitable for supervised learning loss).
\end{itemize}

The VALAN also comes prepackaged with popular methods proposed for navigation on R2R dataset such as pretraining on auxiliary tasks~\cite{Huang2019TransferableRL}. Figure~\ref{fig:r2r_case_study} contains an image of the dashboard depicting validation metrics computed during training a navigation agent on R2R dataset.